\documentclass[runningheads]{llncs}

\usepackage{eccv}

\usepackage[dvipsnames,table]{xcolor}  
\usepackage{booktabs}  
\usepackage{xcolor}    
\usepackage{makecell}  
\usepackage{multirow} 
\usepackage{amsmath}
\usepackage{amsfonts}
\usepackage{amsmath,amssymb}
\usepackage{mathrsfs} 
\usepackage{wrapfig}
\usepackage{caption}
\usepackage{eccvabbrv}

\usepackage{graphicx}
\usepackage{booktabs}

\usepackage[accsupp]{axessibility}  

\usepackage{hyperref}

\usepackage{orcidlink}
\usepackage{marvosym}

\begin{document}

\title{\vspace*{-1.2cm}GeoDiffMM: Geometry-Guided Conditional Diffusion for Motion Magnification}

\titlerunning{GeoDiffMM}

\author{Xuedeng Liu\thanks{Equal contribution. \textsuperscript{\Letter} Corresponding author.} \and
Jiabao Guo\textsuperscript{*} \and
Zheng Zhang \and
Fei Wang \and
Zhi Liu \and
Dan Guo\textsuperscript{\Letter}}

\authorrunning{X.~Liu et al.}

\institute{School of Computer Science and Information Engineering, Hefei University of Technology, China\\
}

\maketitle

\begin{abstract}
Video Motion Magnification (VMM) amplifies subtle macroscopic motions to a perceptible level. Recently, existing mainstream Eulerian approaches address amplification-induced noise via decoupling representation learning such as texture, shape and frequency schemes, but they still struggle to mitigate the interference of photon noise on true micro-motion when motion displacements are very small. We propose \textbf{GeoDiffMM}, a novel diffusion-based Lagrangian VMM framework conditioned on optical flow as a geometric cue, enabling structurally consistent motion magnification. Specifically, we design a Noise-Free Optical Flow Augmentation strategy that synthesizes diverse nonrigid motion fields without photon noise as supervision, helping the model learn more accurate geometry-aware optical flow and generalize better. Next, we develop a Diffusion Motion Magnifier that conditions the denoising process on (i) optical flow as a geometry prior and (ii) a learnable magnification factor controlling magnitude, thereby selectively amplifying motion components consistent with scene semantics and structure. 
Finally, we perform Flow-based Video Synthesis to map the amplified motion back to the image domain with high fidelity. Extensive experiments on real and synthetic datasets show that GeoDiffMM outperforms state-of-the-art methods and significantly improves motion magnification.
\end{abstract}

\section{Introduction}
\label{sec:intro}


Video motion magnification (VMM) makes imperceptible motions visible and continues to draw active research interest~\cite{liu2005motion,Rubinstein13Revealing,le2019seeing}.  Serving as a “motion microscope” , VMM supports a broad range of applications, including micro-expression recognition~\cite{xia2020revealing,wei2022novel,liu2025ammsm}, modal analysis~\cite{eitner2021effect}, and industrial inspection~\cite{perrot2018video,kim2021manifestation}. Because target motions are extremely small, VMM often operates near the photon-noise scale, where noise-driven and motion-driven intensity changes exhibit similar amplitude and statistics. This leads to confusion between noise and true motion, which biases the magnification process and produces undesired motion~\cite{oh2018learning}.

Existing techniques are typically categorized as Eulerian~\cite{Rubinstein13Revealing,wadhwa2013phase,wu2012eulerian,zhang2017video,takeda2022bilateral,wang2024eulermormer,wang2024frequency} or Lagrangian~\cite{flotho2023lagrangian,liu2005motion}. Eulerian methods magnify motion by decomposing local intensity signals in the pixel or frequency domain (\cref{Comparison with existing work}(a)). Despite mitigating amplification noise to some extent, localized analysis remains sensitive to photon

\begin{wrapfigure}[14]{r}{0.5\textwidth} 
\centering
\includegraphics[width=\linewidth]{figure/Fig1.jpg}
\vspace{-.2in}
\caption{\textbf{Comparison with existing VMM methods.} (a) Mainstream Eulerian Method~\cite{oh2018learning,wang2024eulermormer,byung2024learning}. (b) Our GeoDiffMM substantially reduces undesired motion accumulation and enhances the stability of motion magnification.}
\label{Comparison with existing work}
\end{wrapfigure}

\noindent noise and lacks global geometric constraints, often leading to spatial inconsistency and image blur in weak or repetitive textures~\cite{oh2018learning,wang2024eulermormer}. 
Lagrangian methods achieve magnification by explicitly tracking pixel motion trajectories through optical flow. However, existing optical flow methods~\cite{flotho2023lagrangian,liu2005motion} are scenario-dependent and require extensive hyperparameter tuning. Therefore, constructing a geometry-aware magnification paradigm in Lagrangian mode is an important aspect for VMM.


With this motivation, we shift the focus of motion magnification from the pixel domain to  the geometry domain (\cref{Comparison with existing work}(b)). The core design explicitly encodes where motion occurs and how it should be magnified to constrain the solution space and lower the risk of spurious motion or texture-induced bias. In parallel, we adopt conditional generation instead of simple linear magnification, enabling the model to generate  magnified trajectories under real-motion priors. 

In this work, we propose a pioneering attempt, the first diffusion-based 
Lagrangian approach, the Geometry-Guided Conditional Diffusion Motion Magnification (\textbf{GeoDiffMM}) that leverages geometric optical-flow cues as conditioning signals to steer diffusion toward structurally consistent motion magnification. This is a novel attempt to bridge geometric priors with conditional diffusion for motion magnification. We first design a Noise‑free Optical Flow Augmentation (NOFA), which synthesizes diverse nonrigid motion fields without photon noise for supervision. This enables the model to learn more accurate geometry‑aware optical flow and improves generalization. Next, a Diffusion Motion Magnifier (DMM) conditions the denoising process on optical flow as a geometric prior and uses a learnable magnification factor to control the magnitude, thereby selecting motion consistent with scene semantics and structure. Finally, Flow‑based Video Synthesis (FVS) maps the amplified motion back to the image domain with high fidelity. This approach significantly reduces undesired motion and improves motion magnification stability. Our key contributions include:
\begin{itemize}
\item We propose GeoDiffMM, a novel geometry-guided conditional diffusion framework for VMM, representing a pioneering effort in this area.


\item We design Noise-Free Optical Flow Augmentation, which synthesizes diverse nonrigid motion fields without photon noise as supervision, helping the model learn accurate geometry-aware optical flow. 

\item We propose a Diffusion Motion Magnifier that conditions diffusion on optical flow as a geometric prior and controls magnification with a learnable factor, amplifying structure‑consistent motion.

\item Qualitative and quantitative experiments demonstrate that GeoDiffMM significantly improves motion magnification quality over SOTA methods.
\end{itemize}

\section{Related work}
\label{sec:Related_Works}

{\flushleft \textbf{Video Motion Magnification.}}
VMM methods are divided into two paradigms: Eulerian~\cite{zhang2017video,takeda2018jerk,Rubinstein13Revealing,takeda2019video,wadhwa2013phase,wu2012eulerian,takeda2022bilateral} and Lagrangian~\cite{flotho2023lagrangian,liu2005motion}. 
The Eulerian approach captures changes within a fixed region without tracking pixel motion trajectories. In contrast, Lagrangian methods~\cite{liu2005motion,flotho2023lagrangian} achieve magnification by explicitly tracking pixel motion trajectories through optical flow. However, these methods are scenario-dependent, prone to occlusion, and require extensive hyperparameter tuning. Deep learning-based motion magnification methods~\cite{oh2018learning,dorkenwald2020unsupervised,brattoli2021unsupervised,singh2023multi,singh2023lightweight,byung2024learning} have made significant progress in overcoming these issues, focusing on the Eulerian perspective. 
These methods~\cite{singh2023multi,singh2023lightweight,wang2024eulermormer,byung2024learning,wang2024frequency} use representation learning to decouple the input into shape and texture, and represent motion as the inter-frame shape difference between two frames, the input into motion features represented by low frequencies and texture features represented by high frequencies. 
Chen et al.~\cite{chen2023novel} proposed an Eulerian optical flow method that learns  shape and texture and uses shape flow for supervision. However, representation decoupling can blur motion and static features. We address this with a Lagrangian optical‑flow‑conditioned diffusion model reducing manual intervention in VMM.

{\flushleft \textbf{Optical Flow meets Diffusion Model.}}
By modeling spatial–temporal dependencies, diffusion models deliver high-quality images and are standard for video generation~\cite{ho2022video, bar2024lumiere, chen2024videocrafter2, geng2025motion, jin2025flovd, stracke2025cleandift}. 
Recent work~\cite{koroglu2025onlyflow,luo2024flowdiffuser,lew2025disentangled,pan2023self} couples diffusion with optical flow, yielding improved flow estimation. We leverage diffusion to improve optical flow for subtle motions and to mitigate motion blur and photon noise, enhancing video quality. Building on this, we design a Lagrangian flow‑conditioned diffusion paradigm enabling high-quality magnification of subtle video motions.

\section{Methodology}
\subsection{Preliminary}

{\flushleft \textbf{Task Definition.}} Lagrangian motion magnification aims to reveal subtle or imperceptible motions in videos by tracking and modifying pixel trajectories over time. 
Let $I(x, t)$ be the observed video frame at spatial position $x$ and time $t$ is $I(x, t) = f(x + \boldsymbol{\delta}(x, t))$, where $\boldsymbol{\delta}(x, t)$ is the motion field and $f(\cdot)$ maps displacement to intensity~\cite{oh2018learning,wang2024eulermormer}. We visualize motion by amplifying displacement using a magnification factor $\alpha$, i.e., $I_{m}(x, t) = f(x + (1+\alpha)\,\boldsymbol{\delta}(x, t))$. By enhancing subtle motions, lagrangian motion magnification technique allows for easier visualization without altering the video’s overall integrity.

\begin{figure*}[tb]
\centering
\includegraphics[width=1.0\linewidth]{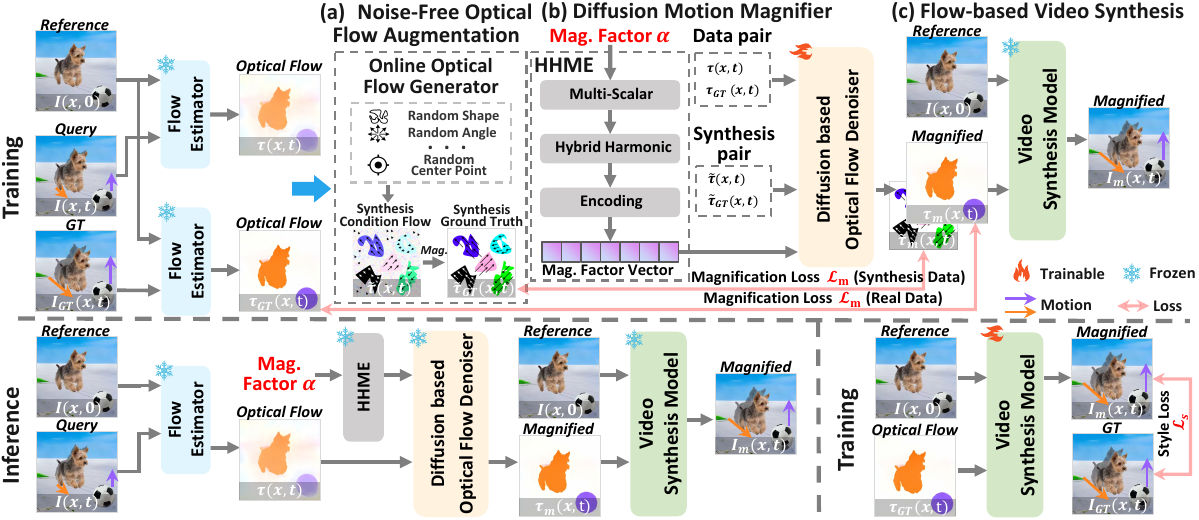}
\caption{\textbf{Overview of the GeoDiffMM pipeline.} (a) Noise-Free optical flow augmentation starts from the optical flow extraction between  reference and  query frame,
structurally stable conditional flow. (b) Diffusion Motion Magnifier is conditioned on the initial optical flow and an encoded magnification factor to produce a structure-consistent magnified flow. (c) Flow-based Video Synthesis uses the reference frame and the magnified flow for pixel-level resampling and reconstruction, producing a high-fidelity magnified frame with less undesired motion.
}
\label{fig:Sec3_overview}
\end{figure*}

{\flushleft \textbf{Overview.}} We aim to model intermediate motion with generative optical flow, enabling a new motion magnification paradigm. As shown in ~\cref{fig:Sec3_overview}, the proposed GeoDiffMM comprises two stages: a motion-field modeling stage and a magnification synthesis stage. The former focuses on constructing and amplifying motion priors, while the latter maps the amplified motion back to the image domain. In the motion field modeling stage, we first use an optical flow estimator~\cite{luo2024flowdiffuser} to extract initial flow from a pair of reference and query frames $\{I(x,0), I(x,t)\}\in\mathbb{R}^{H\times W\times 3}$, serving as an observation of displacement.
At first, we design a \textbf{\textit{Noise-Free Optical Flow Augmentation}} module at the input to generate a structurally reliable, noise-free motion field as synthesis supervisory flow $\tilde{\tau}_{\textnormal{\tiny GT}}(x,t)$, while 
the generated synthetic optical flow along with true optical flow serve complementary data sources and alternate in training. Then, the conditional flows from both the synthesis pair and the real video data pair are fed into a \textbf{\textit{Diffusion Motion Magnifier}}, which is built upon Hybrid Harmonic
Magnification Encoding and Optical Flow Denoiser, semantically important motion
components,
resulting in the motion-magnified flow $\tau_m(x,t)$ with reliable structural consistency constraints. 
This process is faithful to scene structure, preserving intra-region smoothness.
In the magnification synthesis stage, we employ a \textbf{\textit{Flow-based Video Synthesis}} model to drive pixel-level resampling and reconstruction using the magnified flow, producing the magnified frame $I_{m}(x, t)$ at the desired strength. Benefiting from the Noise-free prior and diffusion-based magnification established in the motion-field modeling stage, the magnification synthesis stage reduces undesired motion while enhancing key motion details, leading to more natural and stable visual magnification.

\subsection{Noise-Free Optical Flow Augmentation}

\begin{wrapfigure}[15]{r}{0.5\textwidth} 
\centering
\vspace{-.25in}
\includegraphics[width=\linewidth]{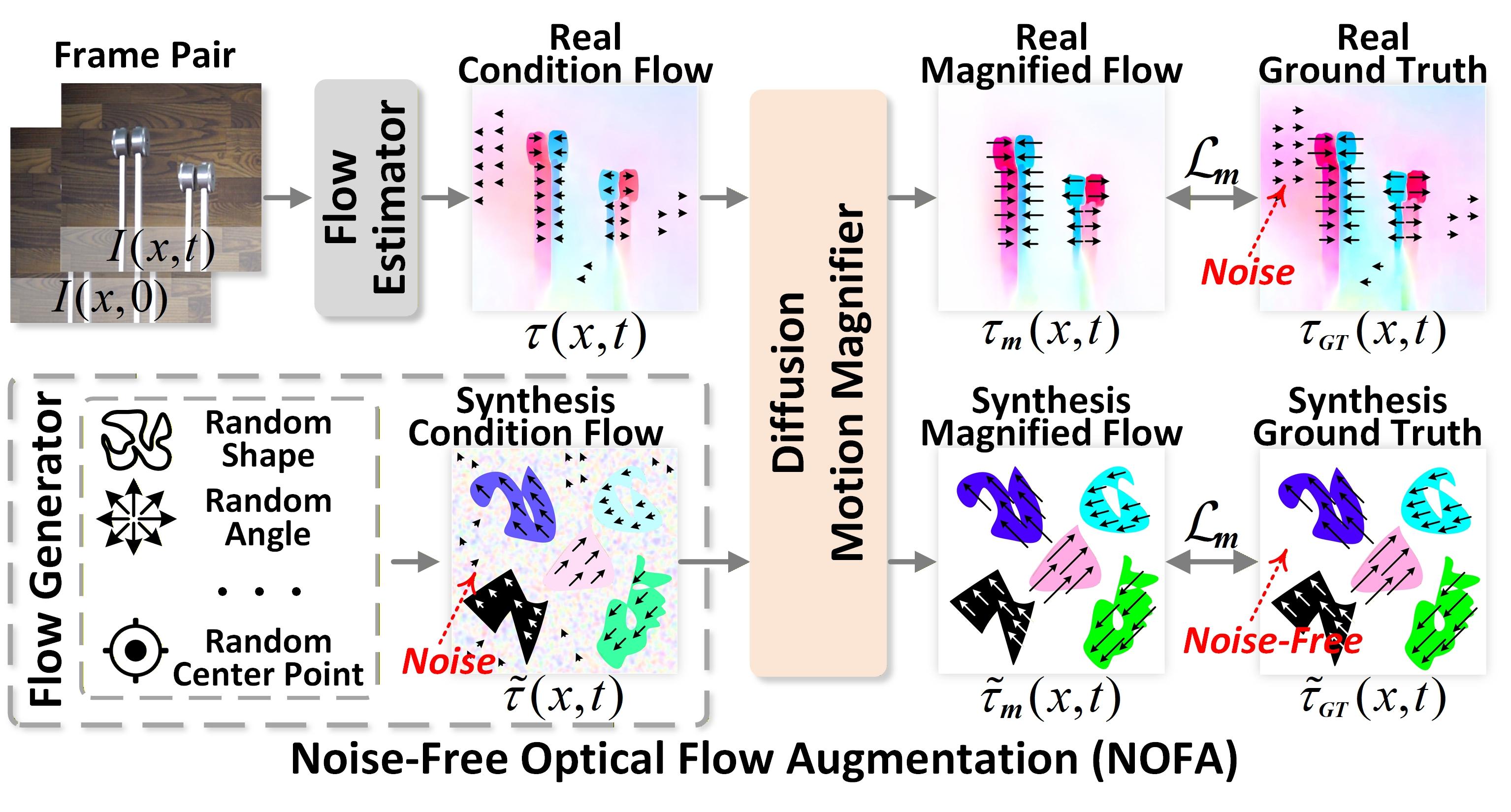}
\vspace{-.3in}
\caption{The proposed \textbf{Noise-Free Optical Flow Augmentation (NOFA).} NOFA composes controllable conditional flow to complement real ones. Both pass through diffusion magnifier to produce magnified flow with synthetic supervision, improving training stability and guiding the model to learn relevant features instead of noise.}
\label{fig:Sec3_NOFA}

\end{wrapfigure}

Existing backend denoising techniques in VMM, whether through filtering~\cite{wang2024frequency,wang2024eulermormer,singh2023lightweight} or regularization~\cite{byung2024learning}, fall short against undesired motion and image blur caused by photon noise and sensor jitter~\cite{oh2018learning}. In order to achieve source-level controllability in optical-flow–based motion field modeling, we present Noise-Free Optical Flow Augmentation (NOFA). As shown in \cref{fig:Sec3_NOFA}, NOFA augment optical flow components to increase directional and morphological diversity. This augmentation improves coverage of motion patterns, leading to more stable training and better generalization.

We first generate arbitrarily shaped masks that support the synthetic conditional optical flow $\tilde{\tau}(x,t)$. Let $\{c_i\}_{i=1}^n$ be $n$ sampled center points serving as spatial anchors. For each region $i$ centered at $c_i$, we select a shape type $\varphi_i$ from the set $\Phi=\{\varphi_{\mathrm{ell}},\varphi_{\mathrm{poly}},\varphi_{\mathrm{frac}},\varphi_{\mathrm{spot}\}}$, corresponding to elliptical, polygonal, fractal-contour and spot structures, then randomly sample its shape parameters such as scale, aspect ratio, orientation and boundary smoothness. Given $\varphi_i$, the sampled parameters, and $c_i$, a mask generator produces a binary mask $M_i \in \{0,1\}^{H\times W}$ by placing the specified shape at $c_i$. The collection $\{M_i\}_{i=1}^n$ provides the spatial support for subsequent flow component synthesis.


Once the masks are defined, the angular range $[0, 2\pi)$ is partitioned into $d$ equal directional segments $l_k = [2\pi k/d, 2\pi(k+1)/d]$. A total of $n$ directions are then uniformly and randomly sampled without replacement from these segments to obtain the direction set $\Theta = \{\theta_1, \dots, \theta_n\}$, ensuring balanced and diverse directional coverage. The direction set $\Theta$ is assigned to each local region: for the $i$-th region, the direction is specified, and the unit direction vector is defined as $[ \cos(\phi_i), \sin(\phi_i)]^T$. For each local region, we define an optical flow magnitude $m$, the conditional optical flow for that region is then $
\tilde{\tau}(x,t) = m \cdot M_i \cdot [\cos(\phi_i), \sin(\phi_i)]^T$.

Based on the conditional flow $\tilde{\tau}(x,t)$, we synthesize the target optical flow $\tilde{\tau}_{\textnormal{\tiny GT}}(x,t)$. A random magnification factor $\alpha$ is applied to generate the corresponding target optical flow $\tilde{\tau}_{\textnormal{\tiny GT}}(x,t)$ for the region by multiplying the conditional optical flow $\tilde{\tau}(x,t)$, as follows:
\begin{equation}
\tilde{\tau}_{\textnormal{\tiny GT}}(x,t) = \alpha \cdot \tilde{\tau}(x,t).
\end{equation}
This linear multiplication method follows industry conventions \cite{oh2018learning,wang2024eulermormer,wang2024frequency,chen2023novel,singh2023lightweight,singh2023multi}, 
which generalizes effectively to capture the noise-free optical flow variation. 

In the regions of the conditional optical flow not covered by the masks, we add noise to these uncovered regions to simulate real-world optical flow, 
with the magnitude drawn from log-normal distribution $\mathrm{LogNormal}(\mu, \sigma)$ and its direction $U(0, 2\pi)$ of the real condition flows that are extracted from query, reference, and ground truth frames. Meanwhile, we apply Gaussian filtering to the noise 
to smooth it and improve spatial continuity, simulating the generated optical flows close to that of real videos. 
As a result, we obtain the smooth-noise conditional optical flow $\tilde{\tau}(x,t)$ and its noise-free target counterpart $\tilde{\tau}{\textnormal{\tiny GT}}(x,t)$. 
More details and experiments about NOFA are provided in the Appendix.

During the model training phase, the synthetic optical flow generated by the Noise-Free Optical Flow Augmentation, along with the true optical flow extracted by the optical flow estimator, serves complementary data sources and alternates in training. Each synthetic batch is generated using independent random seeds, ensuring the diversity of samples in terms of direction, shape, location, and magnitude.

\subsection{Diffusion Motion Magnifier}

To achieve controllable magnification of subtle motions in videos while mitigating the impact of optical flow noise, we propose a Diffusion Motion Magnifier (DMM), which consists of two components: Hybrid Harmonic Magnification Encoding (HHME) and an Optical Flow Denoiser (OFD).

We observe that the magnification factor $\alpha$ is a scalar, which prevents the model from directly perceiving the corresponding amplitude differences and often leads to under-magnified motion. To enable the OFD to accurately capture how motion amplitude varies with different $\alpha$, we introduce a Hybrid Harmonic Magnification Encoding. HHME maps $\alpha$ to a high-dimensional embedding $h_{\alpha}\in\mathbb{R}^{1\times C}$ that is compatible with the diffusion model. Specifically, we first linearly rescale $\alpha$ to $[0,1]$, yielding a rescaled variable $n_\alpha\in\mathbb{R}^{1\times 1}$. This preserves the relative ratios among different $\alpha$ while avoiding training instability due to large numeric ranges. To capture multi-scale variations of $\alpha$, we define frequency scales $f_k=2^k,\ k\in\mathbb{N}$, and apply sine and cosine encodings under each $f_k$ to obtain two variables $c_\alpha\in\mathbb{R}^{1\times k}$ and $s_\alpha\in\mathbb{R}^{1\times k}$, respectively. Finally, we concatenate $n_\alpha$, $c_\alpha$, and $s_\alpha$, and use a fully connected layer $\mathcal{F}_c(\cdot)$ to fuse the multi-frequency harmonic features and adjust the dimensionality, producing a more discriminative embedding $h_\alpha\in\mathbb{R}^{1\times C}$:
\begin{equation}
\begin{aligned} 
c_\alpha &=[\cos(2\pi f_1 n_\alpha)\ \Vert\ \cdots\ \Vert\ \cos(2\pi f_k n_\alpha)]_1, \\
s_\alpha &=[\sin(2\pi f_1 n_\alpha)\ \Vert\ \cdots\ \Vert\ \sin(2\pi f_k n_\alpha)]_1, \\
h_\alpha &=\mathcal{F}_c([\,n_\alpha\ \Vert\ c_\alpha\ \Vert\ s_\alpha\,]_1).
\end{aligned}
\end{equation}

\begin{figure*}[tb]
\centering
\includegraphics[width=0.98\linewidth]{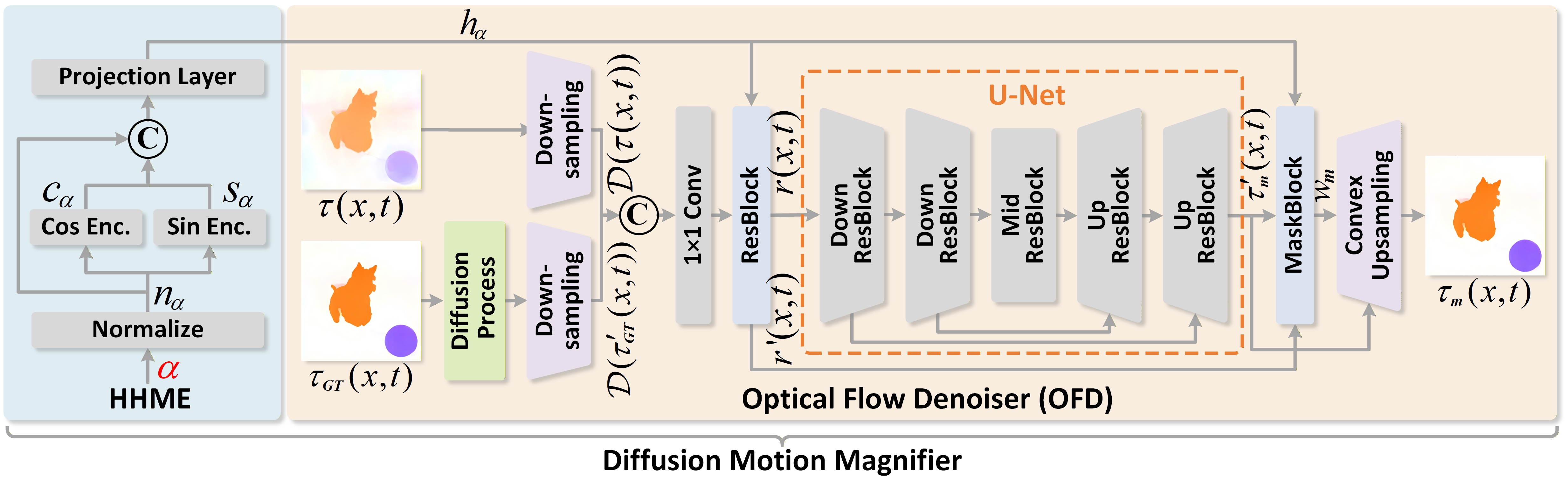}
\caption{Overview of the proposed \textbf{Diffusion Motion Magnifier (DMM)}, illustrating the pipeline from Hybrid Harmonic Magnification Encoding (HHME) to the Optical Flow Denoiser (OFD) and convex upsampling for motion magnification.}
\label{fig:Sec3_DMM}
\end{figure*}

After encoding $\alpha$, we leverage $h_\alpha$ as a condition for the OFD to generate the magnified optical flow, as illustrated in \cref{fig:Sec3_DMM}. We extract the conditional flow $\tau(x,t)\in\mathbb{R}^{H\times W\times 2}$ from the frame pair $\{I(x,0),\, I(x,t)\}\in\mathbb{R}^{H\times W\times 3}$, and the target flow $\tau_{\textnormal{\tiny GT}}(x,t)\in\mathbb{R}^{H\times W\times 2}$ from $\{I(x,0),\, I_{\textnormal{\tiny GT}}(x,t)\}\in\mathbb{R}^{H\times W\times 3}$. The noised version of the target flow under diffusion is denoted by $\tau'_{\textnormal{\tiny GT}}(x,t)\in\mathbb{R}^{H\times W\times 2}$. To extract flow features while reducing computation, we downsample $\tau(x,t)$ and $\tau'_{\textnormal{\tiny GT}}(x,t)$ with a module $\mathcal{D}(\cdot)$, obtaining compact features $\mathcal{D}(\tau(x,t))\in\mathbb{R}^{\frac{H}{8}\times \frac{W}{8}\times C}$ and $\mathcal{D}(\tau'_{\textnormal{\tiny GT}}(x,t))\in\mathbb{R}^{\frac{H}{8}\times \frac{W}{8}\times C}$. The two features are concatenated along channels and lightly fused with a $1{\times}1$ convolution $\mathcal{F}_r(\cdot)$ to yield $f(x,t)\in\mathbb{R}^{\frac{H}{8}\times \frac{W}{8}\times C}$.
\begin{equation}
f(x,t) =\mathcal{F}_r([\,\mathcal{D}(\tau(x,t))\ \Vert\ \mathcal{D}(\tau'_{\textnormal{\tiny GT}}(x,t))\,]_2).
\end{equation}

We combine $f(x,t)$ with $h_\alpha$ and feed them into a Residual Block for initial semantic injection conditioned on $\alpha$ and cross-modal fusion, producing $r(x,t)\in\mathbb{R}^{\frac{H}{8}\times \frac{W}{8}\times C}$ and a residual branch $r'(x,t)\in\mathbb{R}^{\frac{H}{8}\times \frac{W}{8}\times C}$. A U-Net~\cite{ronneberger2015u} $\mathcal{E}(\cdot)$ is applied to perform reverse diffusion on $r(x,t)$, yielding a coarse flow $\tau'_m \in\mathbb{R}^{\frac{H}{8}\times \frac{W}{8}\times C}$:
\begin{equation}
\tau'_m(x,t) =\mathcal{E}(r(x,t)).
\end{equation}

To restore the coarse flow $\tau'_m(x,t)$ to the original resolution while preserving details and maintaining numerical stability, we adopt convex upsampling~\cite{teed2020raft}. Using $r'(x,t)$, $\tau'_m(x,t)$, and $h_\alpha$, the Mask Block $\mathcal{W}(\cdot)$ generates the convex upsampling weights $w_m\in\mathbb{R}^{\frac{H}{8}\times \frac{W}{8}\times (8\times 8\times 9\times 2)}$. Convex upsampling splits each coarse pixel into 8$\times$8 subpixels using weighted 3$\times$3 neighborhood flow with separate component weights. To generate the magnified flow at full resolution, we employ ConvexUpsampling $\mathcal{G}(\cdot,\cdot)$ to merge $\tau'_m(x,t)$ with $w_m$, increasing the resolution from $\frac{H}{8}\times \frac{W}{8}$ to $H\times W$ and producing magnified optical flow $\tau_m(x,t) \in \mathbb{R}^{H\times W\times 2}$: 
\begin{equation}
w_m=\mathcal{W}(\mathcal{F}_w([\,r'(x,t)\ \Vert\ \tau'_m(x,t)\,]_2)+h_\alpha),
\end{equation}
\vspace{-10pt}
\begin{equation}
\tau_m(x,t)=\mathcal{G}(\tau'_m(x,t),\, w_m),
\end{equation}
where $\mathcal{F}_w(\cdot)$ is a $3{\times}3$ convolution used for local context aggregation prior to mask generation.
After computing $\tau_m(x,t)$, we adopt the $\mathcal{L}_1$ norm with $x_0$-prediction~\cite{ho2020denoising} and define the diffusion loss for motion magnification $\mathcal{L}_m$:
\begin{equation}
\mathcal{L}_m = \underset{t \sim U(1,T)}{\mathbb{E}} \big[\lVert \tau_{\textnormal{\tiny GT}}(x,t) - \tau_m(x,t) \rVert_1 \big].
\end{equation}
$\mathcal{L}_m$ provides critical supervision for the model to learn geometry-aware optical flow representations by enforcing consistency between the magnified flow and the ground-truth flow, thereby ensuring that the magnified motion remains consistent with the scene's geometric structure.

\subsection{Flow-based Video Synthesis}
\label{sec:FVS}

To achieve video motion magnification, a synthesis model produces high-fidelity magnified frames conditioned on the input image and optical flow. As a core component of the system, the Video Synthesis Model $\mathcal{S}(\cdot,\cdot)$ critically determines the detail fidelity of magnified frames, and thus warrants targeted optimization. Inspired by prior work~\cite{jin2023unified,lew2025disentangled}, a multi-scale recurrent architecture is designed. With the reference frame $I(x,0)$ and supervisory flow $\tau_{\textnormal{\tiny GT}}(x,t)$, the model generates high-quality magnified frames $I_m(x,t)\in\mathbb{R}^{H\times W\times 3}$, which are concatenated to form a video:
\begin{equation}
I_m(x,t)=\mathcal{S}(I(x,0),\, \tau_{\textnormal{\tiny GT}}(x,t)).
\end{equation}
Losses are computed on the final output $I_m(x,t)$ to improve reconstruction and preserve texture fidelity. The first term is a pixel-wise $\mathcal{L}_1$ loss minimizing the per-pixel discrepancy between $I_m(x,t)$ and the target frame $I_{\textnormal{\tiny GT}}(x,t)$:
\begin{equation}\small
\mathcal{L}_1=\lVert I_{\textnormal{\tiny GT}}(x,t)-I_m(x,t)\rVert_1.
\end{equation}
In addition to the pixel-level $\mathcal{L}_1$ loss, a style loss $\mathcal{L}_G$\cite{gatys2016image} is introduced to enhance style similarity between the generated frame $I_m(x,t)$ and the ground-truth image $I_{\textnormal{\tiny GT}}(x,t)$.  
This loss has been shown to improve the perceptual quality of images~\cite{reda2022film,lew2025disentangled}. The magnified frame $I_m(x,t)$ is fed into a pretrained VGG network, from which a total of $q$ feature layers are extracted, denoted as $F^\ell \in \mathbb{R}^{H_\ell \times W_\ell \times N_\ell}$ for $\ell=1,\dots,q$, where $H_\ell$, $W_\ell$, and $N_\ell$ are the height, width, and number of channels of the $\ell$-th layer. These feature representations together form a feature space that captures texture information. We compute the Gram matrix $G_{ij}^\ell = \langle F_i^\ell, F_j^\ell \rangle \in \mathbb{R}^{N_\ell \times N_\ell}$ to characterize the correlations among feature channels in the $\ell$-th layer, thereby capturing the style representation of the input image, where $F_i^\ell$ and $F_j^\ell$ denote the vectorized feature maps of channels $i$ and $j$, combined via inner product. Similarly, we compute the Gram matrix of the ground-truth image $ I_{\textnormal{\tiny GT}}(x,t)$ as $A^\ell \in \mathbb{R}^{N_\ell \times N_\ell}$. Style loss $\mathcal{L}_G$ is formulated as:
\begin{equation}
\mathcal{L}_G = \sum_{\ell=0}^{q} \frac{w_\ell}{4  H_\ell^2  W_\ell^2 N_\ell^2} \sum_{i=1}^{N_\ell} \sum_{j=1}^{N_\ell} \big(G^\ell_{ij}-A^\ell_{ij}\big)^2,
\end{equation}
where $w_\ell$ is the weighting factor for the $\ell$-th layer.
The final synthesis loss is a weighted combination:
\begin{equation}
\mathcal{L}_s=\lambda_1 \mathcal{L}_1+\lambda_G \mathcal{L}_G,
\end{equation}
with $\lambda_1$ and $\lambda_G$ as their respective weights. Minimizing $\mathcal{L}_s$ optimizes the parameters of the synthesis model, ensuring high-quality magnified images conditioned on optical flow and the reference frame. FVS details are in the Appendix.

{\flushleft \textbf{Training.} }
GeoDiffMM includes a DMM and a Video Synthesis Model, with their training processes as the motion field modeling stage and the magnification synthesis stage, respectively. The DMM is trained on two complementary sources: 1) real optical flow from a video motion magnification dataset~\cite{luo2024flowdiffuser}, and 2) synthetic optical flow generated by NOFA. Real  optical flow are computed from frame pairs $\{I(x,0), I(x,t)\}$ and $\{I(x,0), I_{\textnormal{\tiny GT}}(x,t)\}$ via Flowdiffuser~\cite{luo2024flowdiffuser}, producing conditional $\tau(x,t)$ and supervisory $\tau_{\textnormal{\tiny GT}}(x,t)$. Synthetic optical flow provide the corresponding conditional $\tilde{\tau}(x,t)$ and supervisory $\tilde{\tau}_{\textnormal{\tiny GT}}(x,t)$. Training alternates between the two sources. We apply forward noising to $\tau_{\textnormal{\tiny GT}}(x,t)$ or $\tilde{\tau}_{\textnormal{\tiny GT}}(x,t)$, and condition the DMM on $\tau(x,t)$ or $\tilde{\tau}(x,t)$ together with the magnification factor $\alpha$ to generate the magnified flow $\tau_m(x,t)$. The model is optimized with $\mathcal{L}_m$ via backpropagation until convergence. For the Video Synthesis Model, we use dataset image pairs $\{I(x,0), I_{\textnormal{\tiny GT}}(x,t)\}$ and extract supervisory flow $\tau_{\textnormal{\tiny GT}}(x,t)$ via Flowdiffuser~\cite{luo2024flowdiffuser}, optimizing  for $\mathcal{L}_s$.

{\flushleft \textbf{Inference.}}
Given input frames $\{I(x,0), I(x,t)\}$ and a user-specified magnification factor $\alpha$, the goal is to produce a motion-magnified video. We first compute the conditional flow $\tau(x,t)$ with an optical flow estimator. The factor $\alpha$ is encoded by the HHME to obtain $h_\alpha$, which together with $\tau(x,t)$, conditions the trained Diffusion Motion Magnifier. Starting from random noise, the diffusion process iteratively denoises to generate the magnified flow $\tau_m(x,t)$. Finally, the reference frame $I(x,0)$ and $\tau_m(x,t)$ are fed into the trained Video Synthesis Model, which refines the result through multi-scale processing to produce the magnified image $I_m(x,t)$. Concatenating $I_m(x,t)$ over time yields the final motion-magnified video.

\section{Experiments}
\subsection{Experiments Settings}

{\flushleft \textbf{Datasets.}} {
\textbf{1) Training Dataset.}}
We adopt the motion magnification dataset proposed by LBVMM~\cite{oh2018learning} for training, consistent with existing works~\cite{oh2018learning, singh2023lightweight, singh2023multi, wang2024frequency}. This dataset is synthesized using images from MS COCO~\cite{lin2014microsoft} as backgrounds and objects from PASCAL VOC~\cite{everingham2010pascal} as foregrounds. 
{
\textbf{2) Real-world Test Datasets.}}
As shown in \cref{tab:maniqa}, the real-world dataset comprises twelve distinct motion scenarios~\cite{oh2018learning, singh2023lightweight, singh2023multi, liu2005motion, wu2012eulerian}. Following the protocol in~\cite{byung2024learning}, the evaluation is divided into two modes: \textbf{Static mode}, where inference is performed based on a reference frame $I(x, 0)$ and a query frame $I(x, t)$, suitable for scenarios with subtle motions, and \textbf{Dynamic mode}, where inference is performed based on continuous frames $I(x, t-1)$ and $I(x, t)$, applicable to scenarios involving large-magnitude motions.
{\textbf{3) Synthetic Test Dataset.}}
For precise quantitative evaluation of video motion magnification performance, the field commonly employs synthetic datasets for testing~\cite{byung2024learning, wang2024eulermormer, wang2024frequency, pan2023self, oh2018learning} and cross-dataset performance assessment~\cite{wang2024eulermormer, wang2024frequency, singh2025hierarchical,singh2023lightweight, singh2023multi}. Because the synthetic test split of LBVMM~\cite{oh2018learning} has not been released, we utilize the synthetic test dataset from FD4MM~\cite{wang2024frequency} instead. This dataset is generated following the synthesis principles of LBVMM~\cite{oh2018learning} and Anisotropy~\cite{takeda2019video}. It uses objects from the StickPNG library as foregrounds and images from DIS5K~\cite{qin2022highly} as backgrounds to synthesize ten test videos.

{\flushleft \textbf{Implementation Details.}}
For all experiments, we employ FlowDiffuser~\cite{lew2025disentangled} as the optical flow extractor. The Noise-Free Optical Flow Augmentation is configured with 5 center points $(n = 5)$, 36 directional segments $(d = 36)$, conditional optical flow magnitude $m \in [0, 0.3]$, noise optical flow magnitude sampled from $\mathrm{LogNormal}(-4.303, 0.527)$, and magnification factor $\alpha \in [0, 100]$~\cite{oh2018learning}. The Hybrid Harmonic Magnification Encoding frequency scales parameter $k \in \{1,2,3,4\}$, and the Optical Flow Denoiser has channel dimensions $\{256, 256, 512\}$. During training, the Video Synthesis Model and the Diffusion Motion Magnifier are optimized separately using AdamW~\cite{loshchilovdecoupled}. For diffusion, we adopt a linear noise schedule with $T=1000$ steps and an $x_0$-prediction objective~\cite{ho2020denoising}, and perform 8 sampling steps at inference. The models use a learning rate of $2\times 10^{-4}$, batch size of 4 and mixed precision. 

\subsection{Quantitative Evaluation}

\begin{figure*}[tb]
\centering
\includegraphics[width=1\linewidth]{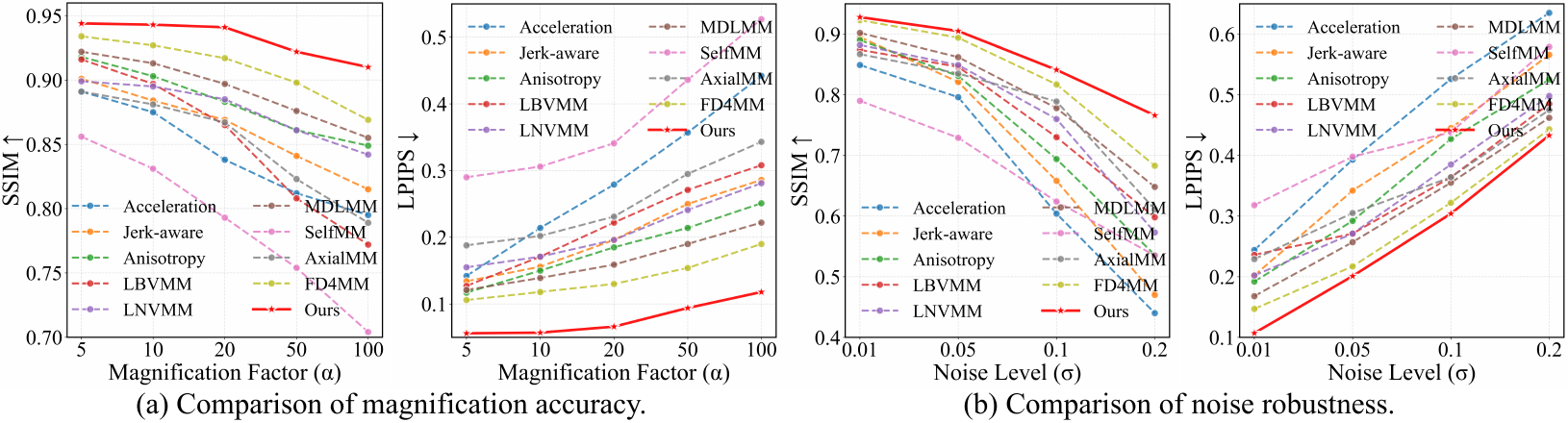}
\caption{\textbf{Performance comparison with state-of-the-art methods on SSIM↑ and LPIPS↓ using the Synthetic Dataset.} (a) Comparison of magnification accuracy for different magnification factor $\alpha$. (b) Evaluation of model robustness under different noise level $\sigma$.}
\label{fig:comparison_four}
\end{figure*}

\begin{figure*}[tb]
\centering
\includegraphics[width=0.98\linewidth]{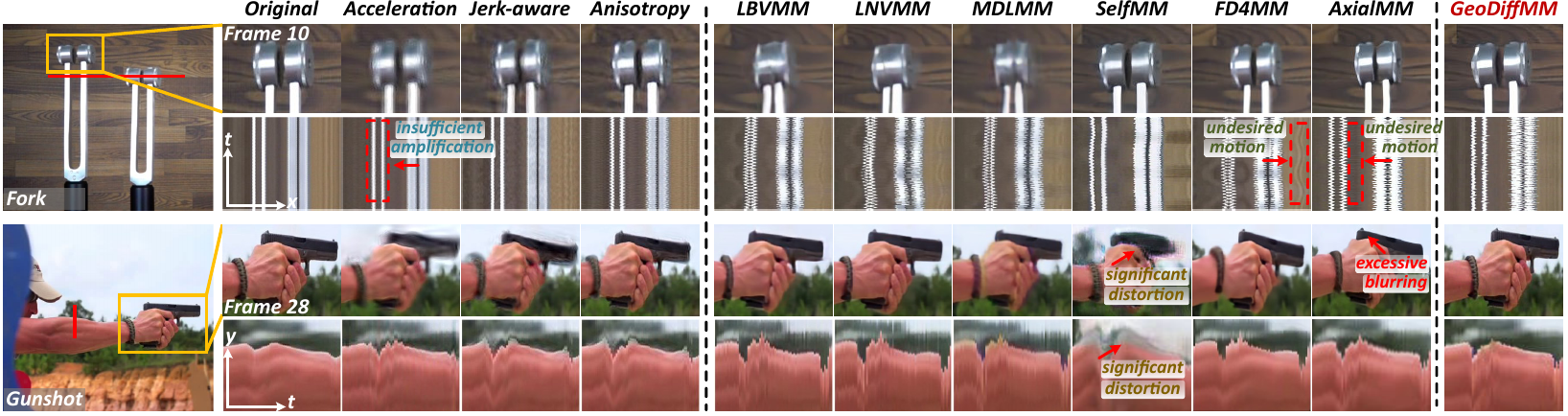}
\vspace{-0.2cm}
\caption{\textbf{Visualization on Real-world Datasets.} We zoom into magnified regions and show spatio-temporal (ST) slices. Hand-crafted filters \cite{zhang2017video,takeda2018jerk,takeda2019video} yield smaller amplification. Self-supervised ~\cite{pan2023self} causes strong distortions. Representation learning ~\cite{oh2018learning,byung2024learning,singh2023lightweight} and frequency decoupling ~\cite{singh2023multi,wang2024frequency} introduce undesired motion and blur. GeoDiffMM achieves the best quality, preserving clarity while avoiding undesired motion. 
}
\label{fig:Visualization}
\end{figure*}

{\textbf{1) Comparison on Real-world Test Datasets.}}
For evaluating motion magnification on real videos, Tab.~\ref{tab:maniqa} presents a comparison of different methods. Anisotropy~\cite{takeda2019video} achieves a high MANIQA score of 0.6535, but it tends to sacrifice motion amplitude in some scenarios (Fig.~\ref{fig:Visualization}). 
Among learning-based methods, SelfMM~\cite{pan2023self} reaches a MANIQA score of 0.6901. 
In contrast, our proposed GeoDiffMM, with a score of 0.7129, significantly surpasses SelfMM~\cite{pan2023self}. 
These results demonstrate that our GeoDiffMM achieves SOTA performance with the highest overall video quality. 
\begin{table*}[tb] 
\caption{\textbf{Quantitative comparison of magnification quality on Real-world Datasets using MANIQA$\uparrow$ scores.} $\alpha$ is 20 in Static Mode and 10 in Dynamic Mode. Our approach achieves the best average score across all tested videos.}
\label{tab:maniqa}
\centering
\resizebox{\linewidth}{!}{
    \begin{tabular}{l | c | c c c c c c | c c c c c c | c}
    \toprule
    \multirow{2}{*}{\textbf{Method}} & \multirow{2}{*}{\textbf{Venue}} 
    & \multicolumn{6}{c|}{\textbf{Static Mode}} 
    & \multicolumn{6}{c|}{\textbf{Dynamic Mode}} 
    & \multirow{2}{*}{\textbf{Avg.}} \\ 
     & & Baby & Fork & Drum & Eng. & Crane & Face & Gun. & Cattoy & Eye & Bottle & Drill & Ball. \\ \midrule

    Acceleration~\cite{zhang2017video} & CVPR'17 & 0.708 & 0.677 & 0.643  & 0.666 & 0.734 & 0.621 & 0.605 & 0.634 & 0.615 & 0.510 & 0.659 & 0.618 & 0.640 \\
    Jerk-Aware~\cite{takeda2018jerk} & CVPR'18 & 0.709 & 0.687  & 0.674 & 0.671 & 0.741 & 0.626 & 0.618  & 0.642 & 0.617 & 0.514 & 0.677 & 0.628 & 0.650 \\ 
    Anisotropy~\cite{takeda2019video} & CVPR'19 & {0.712} & {0.684} & 0.693 & {0.677} & 0.742 & 0.629 & {0.611} & {0.648} & {0.619} & 0.516  & 0.683 & 0.629 & 0.654 \\ \midrule

    LBVMM~\cite{oh2018learning} & ECCV'18 & 0.707 & 0.674 & 0.687 & 0.674  & 0.753 & 0.632 & 0.616  & 0.642 & 0.616 & 0.513 & 0.702  & 0.628 & 0.654 \\
    LNVMM~\cite{singh2023lightweight} & WACV'23 & 0.672 & 0.674 & {0.695} & 0.644 & {0.767} & 0.652 & 0.609  & 0.641 & 0.607  & 0.508  & 0.715 & 0.621 & 0.651 \\
    MDLMM~\cite{singh2023multi} & CVPR'23 & 0.657 & 0.682 & 0.695 & 0.656 & 0.764 & {0.656} & 0.609 & 0.639 & 0.611 & {0.520} & {0.718} & {0.629} & {0.655} \\
    SelfMM~\cite{pan2023self} & NeurIPS'23 & \underline{0.746} & 0.709 & \underline{0.729} & 0.723 & 0.763 & 0.659 & 0.624 & 0.674 & \underline{0.661} & 0.604 & 0.699 & \underline{0.689} & \underline{0.690} \\
    FD4MM~\cite{wang2024frequency} & CVPR'24 & 0.733 & \underline{0.713} & 0.708 & {0.694} & \underline{0.772} & {0.663} & 0.635 & \underline{0.697} & 0.621 & {0.542} & \underline{0.719} & {0.642} & {0.678} \\
    AxialMM~\cite{byung2024learning} & ECCV'24 & 0.731 & 0.673 & 0.717 & \underline{0.736} & 0.758 & \underline{0.674} & \underline{0.638} & 0.673 & 0.657 & \underline{0.635} & 0.647 & 0.654 & 0.682 \\ \midrule

    \rowcolor{gray!20}
    \textbf{GeoDiffMM} & \textbf{-} & \textbf{0.748} & \textbf{0.717} & \textbf{0.757} & \textbf{0.758} & \textbf{0.785} & \textbf{0.683} & \textbf{0.644} & \textbf{0.703} & \textbf{0.686} & \textbf{0.648} & \textbf{0.722} & \textbf{0.704} & \textbf{0.713} \\
    \bottomrule
    \end{tabular}
}
\end{table*}
{\textbf{2) Comparison on Synthetic Test Datasets.}}
We conduct a systematic comparison between our GeoDiffMM method and existing methods~\cite{wang2024frequency, pan2023self, singh2023multi, oh2018learning, zhang2017video, elgharib2015video, takeda2019video}. The experiments cover all parameter settings in the synthetic test dataset, where the motion magnification factor $\alpha \in \{5, 10, 20, 50, 100\}$ and noise level $\sigma \in \{0.01, 0.05, 0.1, 0.2\}$. Quantitative results in \cref{fig:comparison_four} show that GeoDiffMM outperforms competing methods in SSIM and LPIPS.
{\textbf{3) Computational Efficiency Analysis.}}
We evaluate inference efficiency on a single $512 \times 512$ frame using one NVIDIA RTX 5090 GPU. Compared with traditional architectures~\cite{wang2024frequency,pan2023self,byung2024learning}, GeoDiffMM delivers substantial quality gains with only marginal overhead, achieving a 43.5\% LPIPS improvement (\cref{tab:two_stage_vs_end_to_end}). For a comprehensive comparison, we adapt and retrain several representative diffusion models for motion magnification. As these methods were originally designed for general video editing or generation, we align training settings, tune hyperparameters, and integrate HHME to encode the magnification factor for fair evaluation. Their inferior results indicate that generic diffusion architectures lack motion-specific inductive biases for subtle amplification. In contrast, geometry guidance enables GeoDiffMM to outperform diffusion models without optical flow~\cite{yin2025slow,tu2024motioneditor}, and remain superior to flow-based diffusion approaches~\cite{ni2023conditional,lew2025disentangled,jin2025flovd} which also adopt a two-stage design like our framework in both efficiency and generation quality, validating the effectiveness of our design.

\begin{table}[tb]
    \setlength{\abovecaptionskip}{0pt} 
    \setlength{\tabcolsep}{1.5 pt} 

    \caption{Efficiency and quality comparison of different architectures. SSIM/LPIPS are evaluated on the Synthetic dataset, and efficiency is measured per $512\times512$ frame on a single NVIDIA RTX 5090 GPU. Diffusion baselines are retrained for fairness.}
    \centering
    \resizebox{\linewidth}{!}{
        \begin{tabular}{@{}c|c|c|c|c|c|c|c|c@{}}
        \toprule
        Methods &Venue&  Arch. & Flow & Time(s)$\downarrow$ & FLOPs(G)$\downarrow$ & Params(M)$\downarrow$ & SSIM$\uparrow$ & LPIPS$\downarrow$ \\ 
        \midrule
        SelfMM~\cite{pan2023self} &NeurIPS'23&  CNN & $\checkmark$ & 0.8 & 332.1 & 17.29 & 0.788 & 0.380 \\
        AxialMM~\cite{byung2024learning} &ECCV'24& CNN & $\times$ & 0.4 & 169.7 & 0.98 & 0.850 & 0.252 \\
        FD4MM~\cite{wang2024frequency} &CVPR'24&  Transformer & $\times$ & 0.3 & 127.6 & 1.47 & \underline{0.910} & \underline{0.138} \\
        \midrule
        MotionDiff~\cite{tu2024motioneditor} & CVPR'24&Diffusion & $\times$ & 1500.7 & $6.2\! \times\! 10^5$ & $2.1\! \times\! 10^3$ & 0.714 & 0.301 \\
        CausVid~\cite{yin2025slow} & CVPR'25&Diffusion & $\times$ & 6.9 & $2.9\! \times\! 10^3$ & $1.4\! \times\! 10^3$ & 0.768 & 0.256 \\
        LFDM~\cite{ni2023conditional} & CVPR'23& Diffusion & \checkmark & 230.3 & $9.5\! \times\! 10^4$ & 108.74 & 0.629 & 0.491 \\
        FloVD~\cite{jin2025flovd} & CVPR'25&Diffusion & \checkmark & 38.9 & $1.6\! \times\! 10^4$ & $2.2\! \times\! 10^3$ & 0.732 & 0.268 \\
        MoMo~\cite{lew2025disentangled} &AAAI'25& Diffusion & \checkmark & 4.7 & $1.9\! \times\! 10^3$ & 74.67 & 0.785 & 0.243 \\
        \rowcolor{gray!20}
        \textbf{GeoDiffMM}  & \textbf{-} & \textbf{Diffusion} & \textbf{\checkmark} & \textbf{0.8} & \textbf{338.4} & \textbf{76.34} & \textbf{0.932} & \textbf{0.078} \\
        \bottomrule
        \end{tabular}
    }
    \label{tab:two_stage_vs_end_to_end}
\end{table}

\subsection{Ablation Studies}

{\flushleft \textbf{Different Optical Flow Extractor.}}
To assess the impact of the optical flow extractor on motion magnification, we evaluate seven mainstream models in Tab.~\ref{tab:flow_comparison}. GeoDiffMM remains robust across different choices, with all variants based on RAFT~\cite{teed2020raft}, StreamFlow~\cite{sun2024streamflow}, DPFlow~\cite{morimitsu2025dpflow}, and FlowDiffuser~\cite{luo2024flowdiffuser} consistently outperforming the state-of-the-art FD4MM~\cite{wang2024frequency}. Specifically, ARFlow~\cite{liu2020learning} ($A_1$) is unsupervised, GMFlow~\cite{xu2022gmflow} ($A_2$) formulates flow as global matching, SEA-RAFT~\cite{wang2024sea} ($A_3$) improves RAFT, RAFT~\cite{teed2020raft} ($A_4$) adopts iterative optimization, StreamFlow~\cite{sun2024streamflow} ($A_5$) leverages multi-frame prediction for temporal consistency, DPFlow~\cite{morimitsu2025dpflow} ($A_6$) targets ultra-high resolution with a dual-pyramid design, and FlowDiffuser~\cite{luo2024flowdiffuser} ($A_7$) estimates flow via diffusion. Among them, FlowDiffuser achieves the best performance with 0.932 SSIM and 0.078 LPIPS, and is therefore selected as the default extractor.

\begin{table*}[t] 
    \centering
    \caption{Comprehensive ablation studies of the proposed components on the Synthetic Dataset. The best results are in \textbf{bold} and the second-best results are \underline{underlined}.}
    \label{tab:total_ablations}
    \vspace{-12pt}

    \begin{subtable}[t]{0.48\textwidth}
        \centering
        \caption{Ablation of optical flow extractor.}
        \vspace{-4pt}
        \label{tab:flow_comparison}
        \setlength{\tabcolsep}{5pt}
        \renewcommand{\arraystretch}{1.2}
        \resizebox{\linewidth}{!}{ 
            \begin{tabular}{l|l|c|cc}
                \toprule
                & \textbf{Method} &\textbf{Venue} &\textbf{SSIM} $\uparrow$ & \textbf{LPIPS} $\downarrow$ \\
                \midrule
                $A_1$ & ARFlow~\cite{liu2020learning} &CVPR'20& 0.769 & 0.342 \\
                $A_2$ & GMFlow~\cite{xu2022gmflow} &CVPR'22& 0.876 & 0.137 \\
                $A_3$ & SEA-RAFT~\cite{wang2024sea} &ECCV'24& 0.879 & 0.138 \\
                \rowcolor{gray!20}$A_4$ & RAFT~\cite{teed2020raft} &ECCV'20& 0.927 & 0.086 \\
                \rowcolor{gray!20}$A_5$ & StreamFlow~\cite{teed2020raft} &NeurIPS'24& \underline{0.929} & 0.089 \\
                \rowcolor{gray!20}$A_6$ & DPFlow~\cite{morimitsu2025dpflow} &CVPR'25& 0.917 & \underline{0.085} \\
                \rowcolor{gray!20} $A_7$ & \textbf{FlowDiffuser}~\cite{luo2024flowdiffuser} &CVPR'24 &\textbf{0.932} & \textbf{0.078} \\
                \bottomrule
            \end{tabular}
        }
    \end{subtable}
    \hfill
    \begin{subtable}[t]{0.48\textwidth}
        \centering
        \caption{Ablation of NOFA component setup.}
        \vspace{-4pt}
        \label{tab:NOFA}
        \setlength{\tabcolsep}{3pt} 
        \resizebox{\linewidth}{!}{ 
            \begin{tabular}{l | c | c c | c c | c | c}
            \toprule
            & \multirow{2}{*}{\textbf{Noise}}
            & \multicolumn{2}{c|}{\textbf{Direction}} & \multicolumn{2}{c|}{\textbf{Shape}} & \multirow{2}{*}{\textbf{SSIM} $\uparrow$} & \multirow{2}{*}{\textbf{LPIPS} $\downarrow$} \\
            & 
            & \textbf{Single} & \textbf{Multi} & \textbf{Single} & \textbf{Multi} & & \\ 
            \midrule
            $B_1$ & - & - & - & - & - & 0.930 & 0.081 \\
            $B_2$ & - & - &\checkmark & - &\checkmark & \underline{0.931} & \underline{0.080} \\
            \midrule
            $B_3$  & \checkmark & \checkmark & - & \checkmark & - & 0.927 & 0.084 \\
            $B_4$ & \checkmark & \checkmark & - & - & \checkmark & 0.929& 0.084 \\
            \midrule
            $B_5$ & \checkmark & - & \checkmark & \checkmark & - & 0.927 & 0.085 \\
            \rowcolor{gray!20} 
            $B_6$ & \checkmark & \textbf{-} & \textbf{\checkmark} & \textbf{-} & \textbf{\checkmark} & \textbf{0.932} & \textbf{0.078} \\
            \bottomrule
            \end{tabular}
        }
    \end{subtable}
  
    \vspace{8pt} 

    \begin{subtable}[t]{0.48\textwidth}
        \centering
        \caption{Ablation of HHME module.}
        \vspace{-4pt}
        \label{tab:HHME}
        \setlength{\tabcolsep}{15pt}
        \resizebox{\linewidth}{!}{ 
            \begin{tabular}{l|l|cc}
                \toprule
                & \textbf{Method} & \textbf{SSIM} $\uparrow$ & \textbf{LPIPS} $\downarrow$ \\
                \midrule
                $C_1$ & Only Sine & \underline{0.925} & \underline{0.085} \\
                $C_2$ & Only Cosine & 0.923 & 0.086 \\
                \midrule
                $C_3$ & PE~\cite{vaswani2017attention} & 0.906 & 0.110 \\
                $C_4$ & MLP~\cite{von2022diffusers} & 0.916 & 0.087 \\
                $C_5$ & SPE~\cite{mildenhall2021nerf} & \underline{0.927} & \underline{0.085} \\
                \rowcolor{gray!20} $C_6$ & \textbf{HHME} & \textbf{0.932} & \textbf{0.078} \\
                \bottomrule
            \end{tabular}
        }
    \end{subtable}
    \hfill
    \begin{subtable}[t]{0.48\textwidth}
        \centering
        \caption{Ablation of DMM with different optical flow-based diffusions.}  
        \vspace{-4pt}
        \label{tab:network_comparison}
        \setlength{\tabcolsep}{3.3pt}
        \resizebox{\linewidth}{!}{ 
            \begin{tabular}{l|l|c|c|cc}
                \toprule
                & \textbf{Method} &\textbf{Venue} & \textbf{Motion Error} & \textbf{SSIM} $\uparrow$ & \textbf{LPIPS} $\downarrow$ \\
                \midrule
                $D_1$ & LFDM~\cite{jin2025flovd}&CVPR'23 &23.439 & 0.634 & 0.484 \\
                $D_2$ & FloVD~\cite{ni2023conditional}&CVPR'25 & 9.657& 0.762 & 0.251 \\
                $D_3$ & MoMo~\cite{lew2025disentangled} &AAAI'25& 7.224 & \underline{0.796} & \underline{0.230} \\
                \rowcolor{gray!20} $D_4$ & \textbf{DMM} &-& 3.087 & \textbf{0.932} & \textbf{0.078} \\
                \bottomrule
            \end{tabular}
        }
    \end{subtable}
    \vspace{8pt} 

    \begin{subtable}[t]{0.48\textwidth}
        \centering
        \caption{Ablation of denoiser backbone (OFD).}
        \vspace{-4pt}
        \label{tab:U-Net}
        \setlength{\tabcolsep}{6pt}
        
        \resizebox{\linewidth}{!}{ 
            \begin{tabular}{l|l|cccc}
            \toprule
            & \multirow{2}{*}{\textbf{Backbone}} & \textbf{FLOPs} & \textbf{Params} & \multirow{2}{*}{\textbf{SSIM} $\uparrow$} & \multirow{2}{*}{\textbf{LPIPS} $\downarrow$} \\
            &                  & \textbf{(G)}   & \textbf{(M)}    &                                          &                                          \\
            \midrule
                $E_1$ & Standard& 1.899 & \underline{71.91} & 0.893 & 0.137 \\
                $E_2$ & Efficient & \underline{1.383} & \textbf{52.18} & \underline{0.919} & \underline{0.092} \\
                \rowcolor{gray!20} $E_3$ & \textbf{OFD } & \textbf{0.251} & 73.67 & \textbf{0.932} & \textbf{0.078} \\
                \bottomrule
            \end{tabular}
        }
    \end{subtable}
    \hfill
    \begin{subtable}[t]{0.48\textwidth}
        \centering
        \caption{Ablation of video synthesis components.}
        \vspace{-4pt}
        \label{tab:pos_encoding_comparison}
        \setlength{\tabcolsep}{10pt}
        \resizebox{\linewidth}{!}{ 
            \begin{tabular}{l|l|cc}
                \toprule
                & \textbf{Components} & \textbf{SSIM} $\uparrow$ & \textbf{LPIPS} $\downarrow$ \\
                \midrule
                $F_1$ & Backward Warping & 0.911 & 0.122 \\
                $F_2$ & +UNet & \underline{0.926} & \underline{0.088} \\
                \rowcolor{gray!20} $F_3$ & \textbf{+Multi-Scale} & \textbf{0.932} & \textbf{0.078} \\
                \bottomrule
            \end{tabular}
        }
    \end{subtable}
\end{table*}

\begin{figure*}[tb]
\centering
\includegraphics[width=0.98\linewidth]{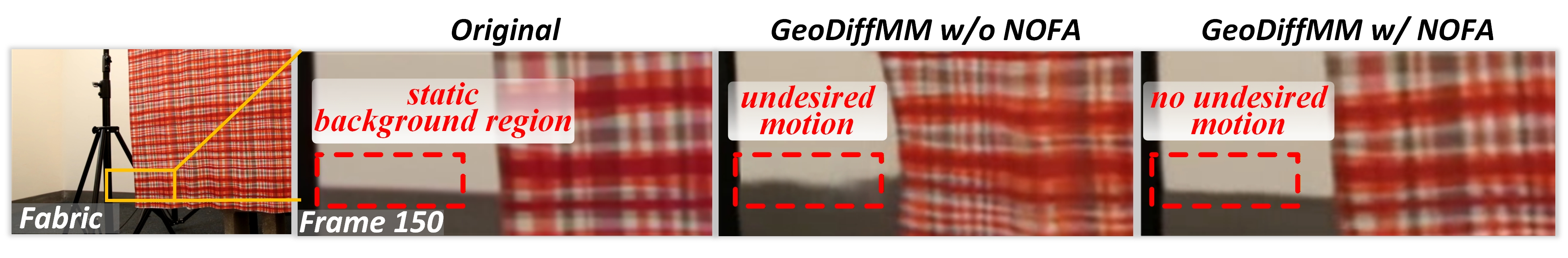} 
\vspace{-0.2cm}
\caption{\textbf{Visual ablation study of NOFA.} Compared with the setting without NOFA ($B_1$), the proposed full NOFA configuration ($B_6$) effectively eliminates undesired motions (see the static background region). This demonstrates that NOFA provides more robust and stable motion magnification.
}
\label{fig:NOFA_}
\end{figure*}

{\flushleft \textbf{Ablation of NOFA Settings.}}
The proposed NOFA synthesizes optical flow with diverse directions and shapes while explicitly injecting noise flow into unmasked regions to mimic real-world disturbances. To evaluate each component, we conduct the ablation study in Tab.~\ref{tab:NOFA}. The baseline without NOFA ($B_1$) achieves 0.930 SSIM and 0.081 LPIPS. Multiple directions and shapes alone ($B_2$) offer limited gains, whereas noise optical flow is crucial. Reduced directional or structural diversity ($B_3$–$B_5$) harms performance, while combining all components ($B_6$) yields the best results of 0.932 SSIM and 0.078 LPIPS. Notably, SSIM~\cite{wang2004image} and LPIPS~\cite{zhang2018unreasonable} are global perceptual metrics and are insensitive to subtle local artifacts caused by noisy flow, often yielding only minor numerical differences. Therefore, quantitative evaluation should be complemented by qualitative comparisons~\cite{wu2012eulerian,oh2018learning,wang2024frequency,wang2024eulermormer}. As shown in Fig.~\ref{fig:NOFA_}, the complete NOFA effectively suppresses undesired motion and produces more stable motion magnification than the baselines.

{\flushleft \textbf{Necessity of HHME.}} 
As a scalar, the motion magnification factor’s encoding significantly affects the model’s perception and the quality of magnification. We compare 3 scalar encoding methods and ablations of our proposed HHME in Tab.~\ref{tab:HHME}. Positional Encoding (PE)~\cite{vaswani2017attention} ($C_3$) is widely used in Transformers to retain positional information. Multilayer Perceptron (MLP)~\cite{von2022diffusers} ($C_4$) encodes scalar labels in diffusion models. Sinusoidal Positional Embedding (SPE)~\cite{pan2023self, mildenhall2021nerf} ($C_5$), used in NeRF~\cite{mildenhall2021nerf} and SelfMM~\cite{pan2023self}, captures high-frequency details.We further ablate HHME using only Sine ($C_1$), only Cosine ($C_2$), and a hybrid of both ($C_6$). HHME ($C_6$) achieves the best SSIM and LPIPS.

{\flushleft \textbf{Ablation of Diffusion Motion Magnifier.}}
To validate DMM, we perform a module-level comparison by fixing the overall pipeline and replacing only the diffusion core (\cref{tab:network_comparison}). Specifically, we adopt the cores from LFDM~\cite{ni2023conditional} ($D_1$), FloVD~\cite{jin2025flovd} ($D_2$), and MoMo~\cite{lew2025disentangled} ($D_3$), which generate conditional optical flow, disentangled camera and object flow, and bidirectional flow for interpolation, respectively. Our DMM ($D_4$) directly takes the magnification factor $\alpha$ and conditional flow $\tau(x,t)$ as inputs. For fairness, we adapt the input layers of $D_1$–$D_3$ to use the same conditions. Since accurate flow generation is central to DMM, we evaluate both image quality (SSIM~\cite{wang2004image}, LPIPS~\cite{zhang2018unreasonable}) and motion error~\cite{pan2023self}, an optical-flow-based metric measuring endpoint error between magnified and ground-truth flow. Within the same framework, DMM achieves 0.932 SSIM, 0.078 LPIPS, and the lowest motion error of 3.087, consistently outperforming $D_1$–$D_3$, demonstrating its superior and task-specific flow generation capability.

\begin{figure*}[tb]
\centering
\includegraphics[width=0.98\linewidth]{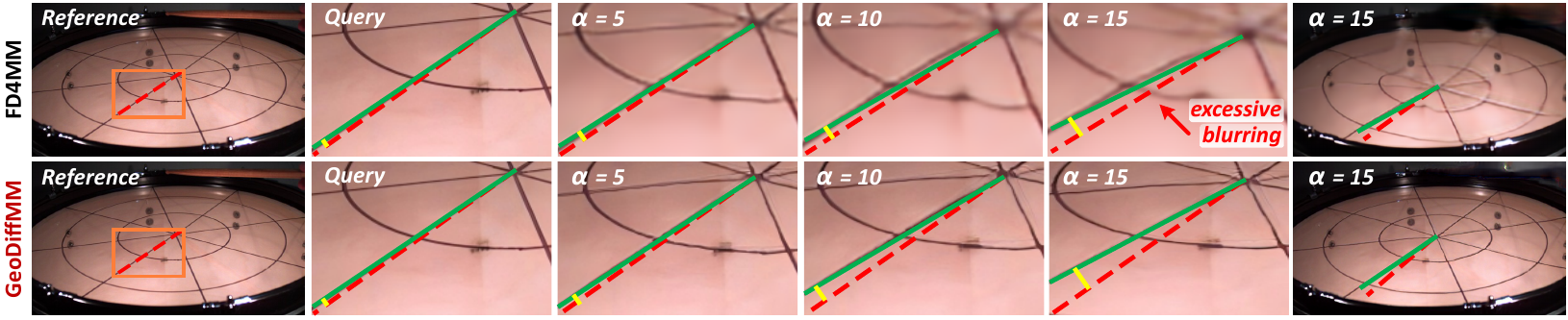}
\vspace{-0.2cm} 
\caption{\textbf{Visualization of motion magnification in static mode for \textit{drum} video from a real-world dataset.} We compare the effects of different magnification factors $\alpha$ on the video’s motion magnification performance.} 
\label{Visualization2}
\end{figure*}

{\flushleft \textbf{Impact of Denoiser Backbone.}} 
We compare our proposed OFD against two baseline diffusion backbones to evaluate its impact on performance in Tab.~\ref{tab:U-Net}. Standard Denoiser~\cite{von2022diffusers} ($E_1$) represents the standard backbone for diffusion models, Efficient Denoiser~\cite{saxena2023surprising} ($E_2$) is the backbone used in the optical flow estimation method DDVM, and OFD ($E_3$) adopted in this work is the convex upsampling diffusion model backbone .
Compared to $E_1$ and $E_2$, OFD ($E_3$) achieves higher SSIM and lower LPIPS while requiring the least FLOPs, highlighting its superior balance of performance and efficiency for Diffusion Motion Magnifier.

{\flushleft \textbf{Ablation of Video Synthesis Model.}} 
To evaluate the contribution of each component in the Video Synthesis Model, we conduct a progressive ablation study by incrementally adding modules (Tab.~\ref{tab:pos_encoding_comparison}).
Backward Warping~\cite{zhou2016view,liu2017video} ($F_1$) uses only the backward warping module. Backward Warping + UNet~\cite{ronneberger2015u} ($F_2$) introduces a U-Net for feature fusion. Based on $F_2$, we incorporate a multi-scale strategy that resizes the reference frame $I_0$ and magnified optical flow $\tau_m(x,t)$ to synthesize the magnified frame $I_m$ in a coarse-to-fine manner.
As components are progressively added, performance improves, with SSIM increasing from 0.911 to 0.932 and LPIPS decreasing from 0.122 to 0.078, demonstrating the effectiveness of both U-Net feature fusion and multi-scale synthesis.
\vspace{-0.3cm}

\subsection{Qualitative Analysis}
{\flushleft \textbf{1) 
Motion Magnification.}}
We zoom into the magnified regions and present the ST slices on the real-world dataset in Fig.~\ref{fig:Visualization}. Hand-crafted methods~\cite{zhang2017video, takeda2018jerk, takeda2019video} fail to amplify subtle motion, and the self-supervised method~\cite{pan2023self} introduces distortions. Representation learning and frequency decoupling approaches generate spurious motion and blur textures in static areas~\cite{byung2024learning, singh2023lightweight, singh2023multi, oh2018learning}. GeoDiffMM achieves the cleanest visual results, preserving sharpness without inducing undesired motion. 
{
\textbf{2) 
Magnification Factor.}}
We evaluate a real-world drum video in static mode with magnification factors $\alpha \in \{5, 10, 15\}$ . As Fig.~\ref{Visualization2} shows, motion magnification increases with $\alpha$, and compared to the method~\cite{wang2024frequency}, GeoDiffMM maintains high-quality video while achieving greater amplification. More ablation experiments and visualizations can be seen in Appendix.

\vspace{-0.2cm}
\section{Conclusion}
We propose GeoDiffMM, a geometry-guided conditional diffusion framework for video motion magnification. By conditioning on optical flow and encoding the magnification factor, our Diffusion Motion Magnifier amplifies structure-consistent motion. Together with Noise-Free Optical Flow Augmentation and a Flow-based Video Synthesis model, GeoDiffMM produces high-fidelity magnified frames with less undesired motion and greater stability. Results on synthetic and real data consistently outperform SOTA, confirming the effectiveness of our Lagrangian geometry-conditioned paradigm.

%
%
\bibliographystyle{splncs04}
\bibliography{main}
\end{document}